\def\year{2019}\relax
\documentclass[letterpaper]{article} 
\usepackage{aaai19}  
\usepackage{times}  
\usepackage{helvet} 
\usepackage{courier}  
\usepackage[hyphens]{url}  
\usepackage{graphicx} 
\def\UrlFont{\rm}  
\usepackage{graphicx}  
\usepackage{tabularx}

\usepackage{microtype}
\usepackage{tabularx}
\usepackage{graphicx}
\usepackage{subfigure}
\usepackage{booktabs} 
\usepackage{amsmath} 
\usepackage[utf8]{inputenc}
\usepackage{textcomp}
\usepackage{url}
\def\UrlBreaks{\do\/\do-}
\usepackage[hang,flushmargin]{footmisc}

\usepackage{hyperref}

\newcommand{\minisection}[1]{\vspace{0.04in} \noindent {\bf #1}\ \ }

\title{DeepCrawl: Deep Reinforcement Learning for Turn-based Strategy Games}
\author{Alessandro Sestini,\textsuperscript{\rm 1} Alexander Kuhnle,\textsuperscript{\rm 2} Andrew D. Bagdanov\textsuperscript{\rm 1}\\ 
\textsuperscript{\rm 1}Dipartimento   di   Ingegneria   dell’Informazione,    Università degli  Studi  di  Firenze, Florence,  Italy\\
\textsuperscript{\rm 2}Department  of  Computer   Science   and   Technology,   University   of   Cambridge, United  Kingdom\\ 
\{alessandro.sestini, andrew.bagdanov\}@unifi.it
}

\begin{document}

\maketitle

\begin{abstract}
	In this paper we introduce DeepCrawl, a fully-playable Roguelike
	prototype for iOS and Android in which all agents are controlled by
	policy networks trained using Deep Reinforcement Learning (DRL). Our
	aim is to understand whether recent advances in DRL can be used to
	develop convincing behavioral models for non-player characters in
	videogames. We begin with an analysis of requirements that such an
	AI system should satisfy in order to be practically applicable in
	video game development, and identify the elements of the DRL model
	used in the DeepCrawl prototype. The successes and limitations of
	DeepCrawl are documented through a series of playability tests
	performed on the final game. We believe that the techniques we
	propose offer insight into innovative new avenues for the
	development of behaviors for non-player characters in video games,
	as they offer the potential to overcome critical issues with
	classical approaches.
\end{abstract}

\section{Introduction}
\label{sec:into}
Technological advances in gaming industry have resulted in the production
of increasingly complex and immersive gaming environments. However,
the creation of Artificial Intelligence (AI) systems that control
non-player characters (NPCs) is still a critical element in the
creative process that affects the quality of finished games. This
problem is often due to the use of classical AI techniques that result in
predictable, static, and not very convincing NPC
strategies. Reinforcement learning (RL) can help overcome these issues
providing an efficient and practical way to define NPC behaviors, but
its real application in production processes has issues that can be
orthogonal to those considered to date in the academic field: how can
we improve the gaming experience? How can we build credible and
enjoyable agents? How can RL improve over classical algorithms for
game AI? How can we build an efficient ML model that is also usable on
all platforms, including mobile systems?

At the same time, recent advances in Deep Reinforcement Learning (DRL)
have shown it is possible to train agents with super-human skills able
to solve a variety of environments. However the main objective of DRL
of this type is training agents to surpass human players in
competitive play in classical games like Go~\cite{silver16} and video
games like DOTA 2~\cite{openai18}. The resulting, however, agents
clearly run the risk of being too strong, of exhibiting artificial
behavior, and in the end not being a \emph{fun} gameplay element.

Video games have become an integral part of our entertainment
experience, and our goal in this work is to demonstrate that DRL
techniques can be used as an effective \emph{game design} tool for
learning compelling and convincing NPC behaviors that are natural,
though not superhuman, while at the same time provide challenging and
enjoyable gameplay experience. As a testbed for this work we developed
the DeepCrawl Roguelike prototype, which is a turn-based strategy game
in which the player must seek to overcome NPC opponents and NPC agents
must learn to prevent the player from succeeding. We emphasize that
our goals are different than those of AlphaGo and similar DRL systems
applied to gameplay: for us it is essential to limit agents so they
are beatable, while at the same time training them to be convincingly
competitive. In the end, playing against superhuman opponents is not
\emph{fun}, and neither is playing against \emph{trivial} ones. This
is the balance we try to strike.

With this paper we propose some requirements that a Machine Learning 
system should satisfy in order 
to be practically applied in videogame
production as an active part of game design.
We also propose an efficient DRL system that is able to create a variety of
NPC behaviors for Roguelike games only by changing some parameters in the training set-up; moreover
the reduced complexity of the system and the particular net architecture 
make it easy to use and to deploy to systems like mobile devices. Finally
we introduce a new game prototype, tested with a series
of playability tests, that can be a future benchmark
for DRL for videogame production.

\section{Related work}
\label{sec:related}

Game AI has been a critical element in video game production since the
dawn of this industry; agents have to be more and more realistic and
intelligent to provide the right challenge and level of enjoyment to
the user. However, as game environments have grown in complexity over
the years, scaling traditional AI solutions like Behavioral Trees (BT)
and Finite State Machines (FSM) for such complex contexts is an open
problem~\cite{yannakakis18}.

Reinforcement Learning (RL)~\cite{sutton98} is directly concerned with
the interaction of agents in an environment. RL methods have been
widely used in many disciplines, such as robotics and operational
research, and games. The breakthrough of applying DRL by DeepMind in
2015~\cite{mnih15} brought techniques from supervised Deep Learning
(such as image classification and Convolutional Neural Networks) to
overcome core problems of classical RL. This combination of RL and
neural networks has led to successful application in games. In the
last few years several researchers have improved upon the results
obtained by DeepMind. For instance, OpenAI researchers showed that
with an Actor Critic~\cite{konda03} algorithm such as Proximal Policy
Optimization (PPO)~\cite{schulman17} it is possible to train agents to
superhuman levels that can win against professional players in complex
and competitive games such as DOTA~2~\cite{openai18} and
StarCraft~\cite{deepmind19}.


As already discussed in the introduction, most of the works in DRL aim
to build agents replacing human players either in old-fashioned games
like Go or chess~\cite{silver16,asperti18} or in more recent games
such as Doom or new mobiles 
games~\cite{openai18,vinyals17,inseok19,kempka16,juliani19}. Our 
objective, however, is not to create a new
AI system with superhuman capabilities, but rather to create ones that
constitute an active part of the game design: our work has a similar
conetxt to \cite{zhao19}, while \cite{amato10} tried to use simple RL to 
build a variety of NPC beahviors.

\section{Game design and desiderata}
\label{sec:design}

\begin{figure}
	\vskip 0.2in
	\begin{center}
		\centerline{\includegraphics[width=0.85\columnwidth]{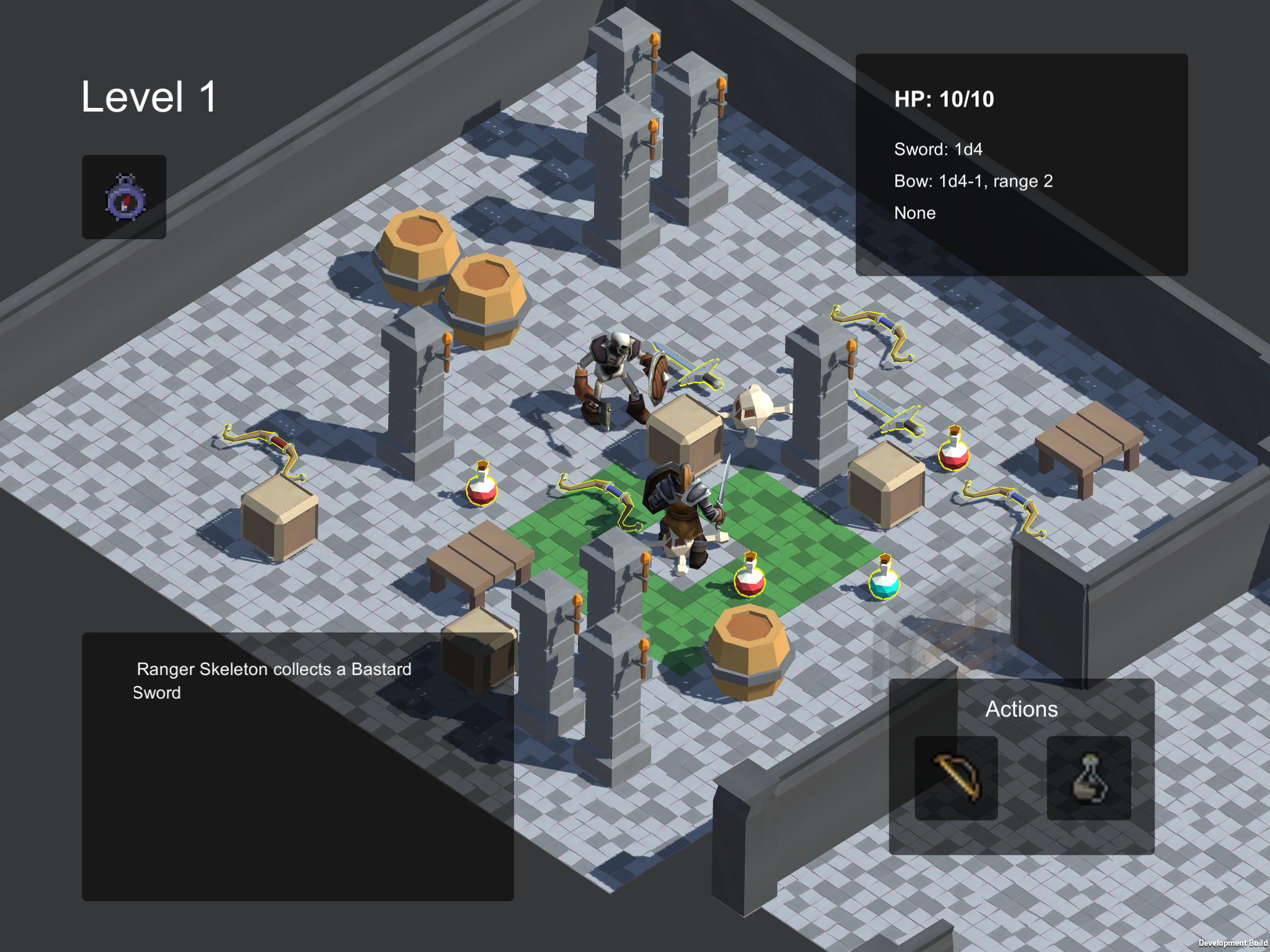}}
		\caption{Screenshot of the final version of DeepCrawl. Each level
			of the game consists of one or more rooms, in each of which
			there is one or more agents. To clear a level the player must fight 
			and win against
			all the enemies in the dungeon. The game is won if the player
			completes ten dungeon levels.}
		\label{screenshot}
	\end{center}
	\vskip -0.2in
\end{figure}

In this section we describe the main gameplay mechanics of DeepCrawl
and the requirements that the system should satisfy in order to be
used in a playable product. The DeepCrawl prototype is a fully
playable Roguelike game and can be downloaded for Android and iOS
\footnote{Android Play: \url{http://tiny.cc/DeepCrawl}
	\\ App Store: \url{http://tiny.cc/DeepCrawlApp}}. In
figure~\ref{screenshot} we give a screenshot of the final game.

\begin{figure*}[ht]
	\vskip 0.2in
	\begin{center}
		\centerline{\includegraphics[width=1.75\columnwidth]{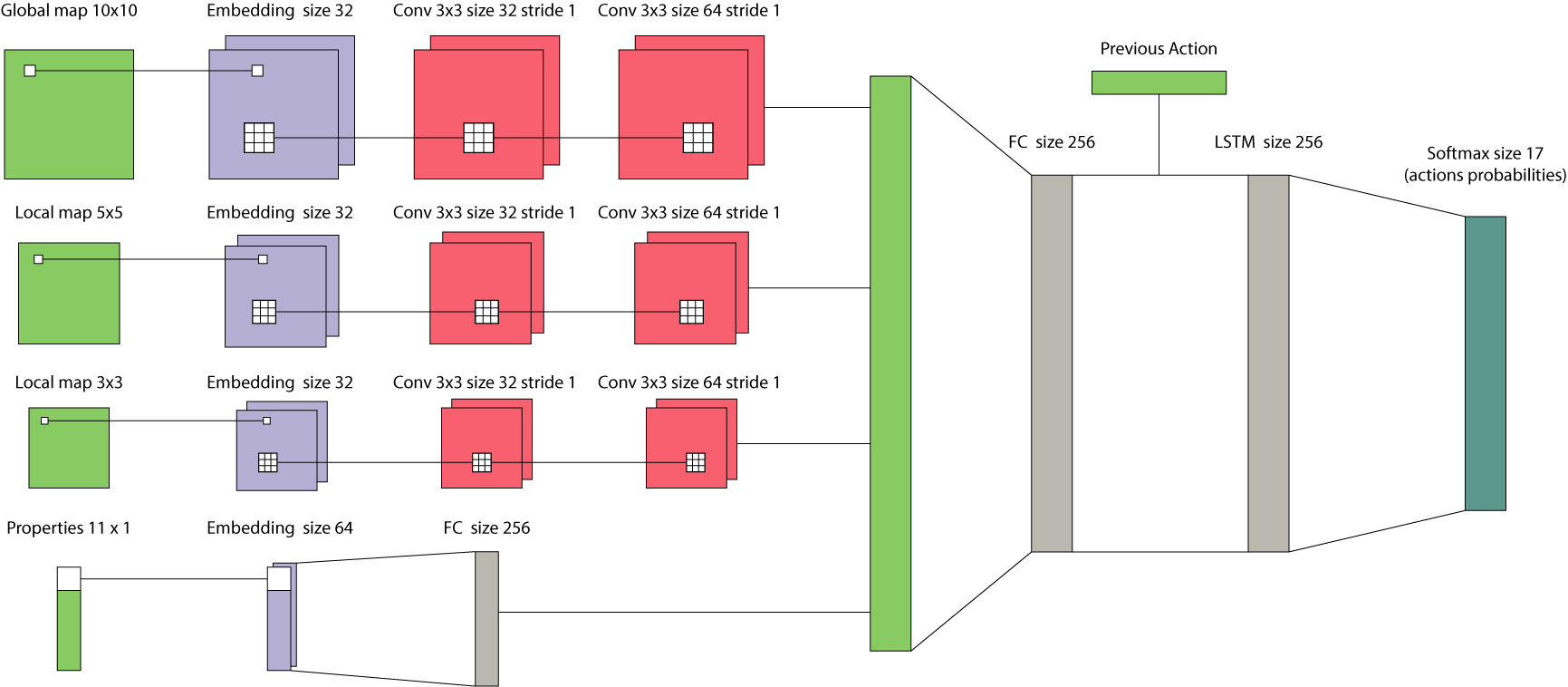}}
		\caption{The policy network used for NPCs in DeepCrawl. See
			section~\ref{architecture} for a detailed description.}
		\label{net}
	\end{center}
	\vskip -0.2in
\end{figure*}

\subsection{Gameplay mechanics}
We decided to build a \emph{Roguelike} as
there are several aspects of this type of games that make them an interesting
testbed for DRL as a game design tool, such as the
\textit{procedurally created environment}, the \textit{turn-based} system
and the \textit{non-modal} characteristic that makes available every action
for actors regardless the level of the game.
In fact, Roguelikes are often used as a proving ground game genre
specifically because they involve a limited set of game mechanics,
which allows game designers to concentrate on emergent complexity of
gameplay as a combination of the relatively simple set of
possibilities given to the player.

The primary gameplay mechanics in DeepCrawl are defined in terms of
several distinct, yet interrelated elements.

\minisection{Actors.} Success and failure in DeepCrawl is based on
direct competition between the player and one or more 
agents guided by a deep policy network trained
using DRL. Player and agents act in procedurally generated
rooms, and player and agents have exactly the same characteristics,
can perform the same actions, and have access to the same information
about the world around them.

\minisection{Environment.}  The environment visible at any instant in
time is represented by a random grid with maximum size of
$10 \times 10$ tiles. Each tile can contain either an agent or player,
an impassible object, or collectible loot. Loot can be of three types:
melee weapons, ranged weapons, or potions. Moreover, player and agent
are aware of a fixed number of personal characteristics such as HP,
ATK, DEX, and DEF. Agents
and player are also aware of their inventory in which loot found on
the map is collected. The inventory can contain at most one object per
type at a time, and a new collected item replaces the previous
one. The whole dungeon is composed of multiple rooms, where in each of
them there are one or more enemies. The range of action of each NPC
agent is limited to the room where it spawned, while the player is
free to move from room to room.

\minisection{Action space.}  Each character can perform 17 different
discrete actions:
\begin{itemize}
	\itemsep0em
	\item \textbf{8 movement actions} in the horizontal, vertical and
	diagonal directions; if the movement ends in a tile containing
	another agent or player, the actor will perform a melee attack: this
	type of assault deals random damage based on the melee weapon
	equipped, the ATK of the attacker, and the DEF of the defender;
	
	\item \textbf{1 use potion action}, which is the only action that does not end
	the actor's turn.  DeepCrawl has two buff potions available, one that increases
	ATK and DEF for a fixed number of turns, and heal
	potion that heals a fixed number of HP; and
	
	\item \textbf{8 ranged attack actions}, one for each possible
	direction. If there is another actor in selected direction, a ranged
	attack is performed using the currently equipped ranged weapon. The
	attack deals a random amount of damage based on the ranged weapon
	equipped, the DEX of the attacker, and the DEF of the defender.
\end{itemize}

\subsection{Desiderata}
\label{desiderata}
As defined above, our goals were to create a playable game, and in order to 
do this the game must be enjoyable from the player's perspective. Therefore, 
in the design phase of this work it was fundamental to define the requirements 
that AI systems controlling NPCs should satisfy in order to be
generally applicable in videogame design:

\minisection{Credibility.}
NPCs  must be \textit{credible}, that is they should act in ways that are 
predictable and that can be interpreted as intelligent. The agents must offer 
the right challenge to the player and should not make counterintuitive moves. 
The user should not notice that he is playing against an AI.

\minisection{Imperfection.} At the same time, the agents must be
\textit{imperfect} because a superhuman agent is not suitable in a
playable product. In early experiments we realized that it was
relatively easy to train \emph{unbeatable} agents that were, frankly,
no fun to play against. It is important that the player always have the
chance to win the game, and thus agents must be beatable.

\minisection{Prior-free.} Enemy agents must be \textit{prior-free} in
that developers do not have to manually specify strategies -- neither
by hard-coding nor by carefully crafting specific rewards -- specific
to the game context prior the training. The system should extrapolate strategies
independently through the trial-and-error mechanism of DRL. Moreover,
this prior-free system should generalize to other Roguelike games
sharing the same general gameplay mechanics.

\minisection{Variety.} It is necessary to have a certain level of
\textit{variety} in the gameplay dynamics. Thus, it is necessary to
support multiple agents during play, each having different
behaviors. The system must provide simple techniques to allow
agents to extrapolate different strategies in the training phase.

\begin{figure*}
	\begin{tabularx}{\textwidth}{XX}
		\includegraphics[width=0.60\textwidth]{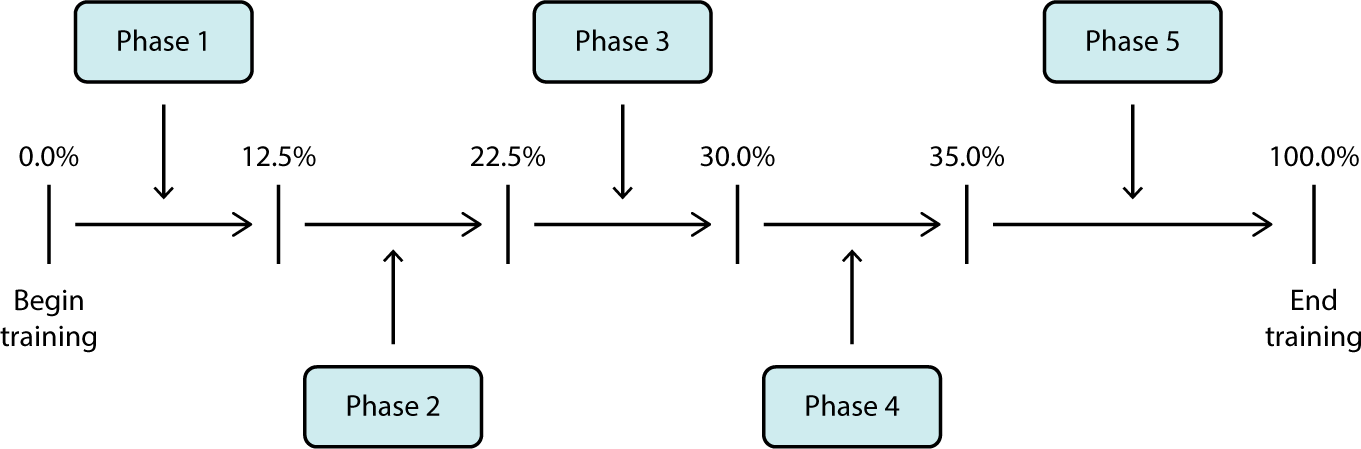} &
		{\hspace{0.8in}\begin{minipage}{0.35\textwidth}
				\resizebox{\textwidth}{!}{%
					\begin{tabular}{c|c|c|c}
						\toprule
						& agent HPs & enemy HP & loot quantity \\ 
						\midrule
						Phase 1  & 20 & 1 & 20\% \\
						Phase 2  &  [5, 20] & 10 & 20\%  \\												
						Phase 3  &  [5, 20] &  [10, 20] & 20\% \\
						Phase 4  &  [5, 20] &  [10, 20] &  [10\%, 20\%] \\
						Phase 5  &  [5, 20] &  [10, 20] &  [5\%, 20\%]\\	
						\bottomrule
				\end{tabular}}
		\end{minipage}}
	\end{tabularx}
	\caption{Curriculum used for training all agents. Left: a training
		timeline showing how long each curriculum phase lasts as a
		percentage of total training steps. Right: the changing generation
		parameters of all the curriculum phases. The numbers in parentheses
		refer to a random number in that; the loot quantity depends on the 
		number of empty tiles in the room (e.g. 20\% loot quantity indicates that
		the number of items in the room is equal to the 20\% of empty tiles in that).
		\label{curriculum_timeline}
	}
\end{figure*}

\section{Proposed model}
\label{sec:model}

Here we describe in detail the main elements of the DRL model that
controls the agent in DeepCrawl, with particular attention to the
neural network architecture and the reward function.

\subsection{Policy network and state representation}
\label{architecture}

We used a policy-based method to learn the best strategy for agents
controlling NPCs. For these methods, the network must approximate the
best policy. The neural network architecture we used to model the
policy for NPC behavior is shown in figure~\ref{net}. The network
consists of four input branches:
\begin{itemize}
	\itemsep0em
	\item the first branch takes as input the whole map of size
	$10 \times 10$, with the discrete map contents encoded as integers:
	\begin{itemize}
		\itemsep0em
		\item 0 = impassable tile or other agent;
		\item 1 = empty tile;
		\item 2 = agent;
		\item 3 = player; and
		\item 4+ = collectible items.
	\end{itemize}
	This input layer is then followed by an embedding layer which
	transforms the $10 \times 10 \times 1$ integer input array into a
	continuous representation of size $10 \times 10 \times 32$, a convolutional 
	layer with 32 filters of size $3 \times 3$, and another $3 \times 3$
	convolutional layer with  64 filters.
	\item The second branch takes as input a local map with size
	$5 \times 5$ centered around the agent's position. The map encoding
	is the same as for the first branch and an embedding layer is
	followed by convolutional layers with the same structure as
	the previous ones.
	\item The third branch is structured like the second, but with a local
	map of size $3 \times 3$.
	\item The final branch takes as input an array of 11 discrete
	values containing information about the agent and the player:
	\begin{itemize}
		\itemsep0em
		\item agent HP in the range [0,20];
		\item the potion currently in the agent's inventory;
		\item the melee weapon currently in the agent's inventory;
		\item ranged weapon in the agent's inventory;
		\item a value indicating whether the agent has an active buff;
		\item a value indicating whether the agent can perform a ranged attack and in which direction;
		\item player HP in the range [0,20];
		\item the potion currently in the player's inventory;
		\item the melee weapon in the player's inventory;
		\item the ranged weapon in the player's inventory; and
		\item a value indicating whether the player has an active buff.
	\end{itemize}
	This layer is followed by an embedding layer of size 64 and a fully-connected (FC) layer of size 256.\\
\end{itemize}
The outputs of all branches are concatenated to form a single vector
which is passed through an FC layer of size 256; we add a
\textit{one-hot} representation of the action taken at the previous
step, and the resulting vector is passed through an LSTM
layer.  The final output of the net is a 
probability distribution over the
action space (like all policy-based methods such as PPO). 

With this model we also propose some novel solutions that have improved
the quality of agents behavior, overcoming some of the challenges
of DRL in real applications:
\begin{itemize}
	\itemsep0em
	\item \textbf{Global vs local view}: we discovered that the use of
	both global and local map representations improves the score
	achieved by the agent and the overall quality of its behavior. The
	combination of the two representations helps the agent evaluate both
	the general situation of the environment and the local details close
	to it; we use only two levels of local maps, but for a more complex
	situation game developers could potentially use more views at different
	scales;
	\item \textbf{Embeddings}: the embedding layers make it possible for
	the network to learn continuous vector representations for the
	meaning of and differences between integer inputs. Of particular
	note is the embedding of the last branch of the network, whose
	inputs have their own ranges distinct from each other, which helps
	the agent distinguish the contents of two equal but semantically
	different integer values. For instance:
	\begin{itemize}
		\itemsep0em
		\item agent HP $\in [0,20]$;
		\item potion $\in [21,23]$;
		\item etc.
	\end{itemize}
	\item \textbf{Sampling the output}: instead of
    taking the action with the highest probability, we sample the output,
    thus randomly taking \textit{one of the most probable actions}. This
    behavior lets the agent make some mistakes during its interaction with
    the environment, guaranteeing \textit{imperfection} and avoids the
    agent getting stuck in repetitive loops of the same moves.
\end{itemize} 


\subsection{Reward function}
When defining the reward function for training policy networks, to
satisfy the \textit{prior-free} requirement we used an extremely
sparse function:
\begin{equation}
R(t)= -0.01 + 
\begin{cases}
-0.1 & \text{for an impossible move} \\
+10.0*\mathrm{HP} & \text{for the win}
\end{cases},
\end{equation}
where HP refers to the normalized agent HPs remaining at the moment
of defeating an opponent. This factor helps the system to learn as
fast as possible the importance of HP: winning with as many HP as
possible is the implicit goal of  a general Roguelike game.

\subsection{Network and training complexity} 
All training was done on an NVIDIA 1050ti GPU with 4GB of RAM. On this
modest GPU configuration, complete training of one agent takes about
two days. However, the reduced size of our policy networks (only about
5.5M parameters in the policy and baseline networks combined) allowed
us to train multiple agents in parallel. Finally, the trained system
needs about 12MB to be stored. We remind though that more agents of
the same type can use the same model: therefore this system does not
scale with the number of enemies, but only with the number of
different classes.

\begin{figure*}
	\begin{tabularx}{\textwidth}{XXX}
		\includegraphics[width=0.32\textwidth]{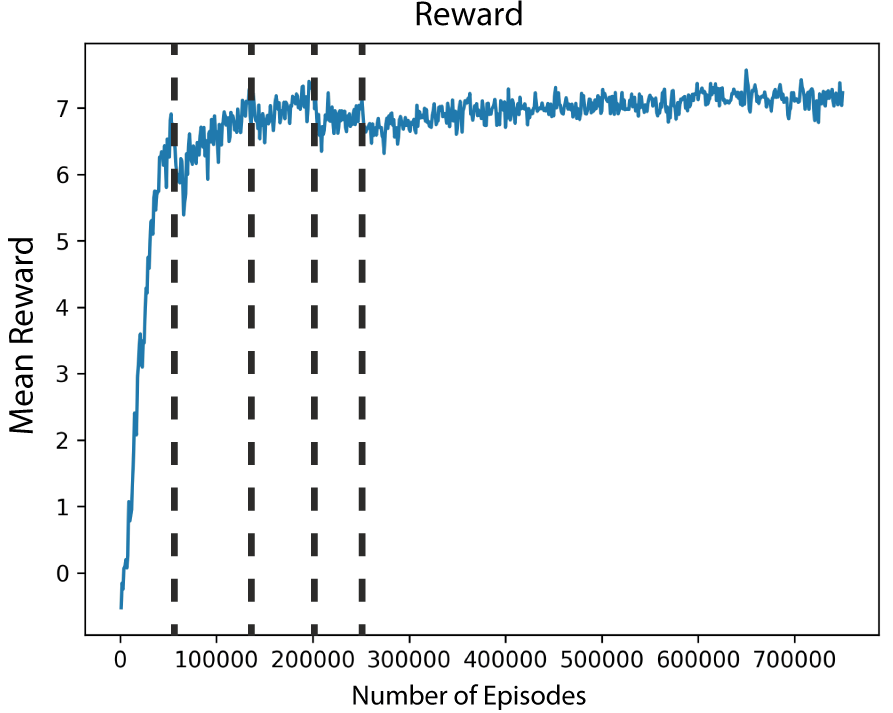} &
		\includegraphics[width=0.32\textwidth]{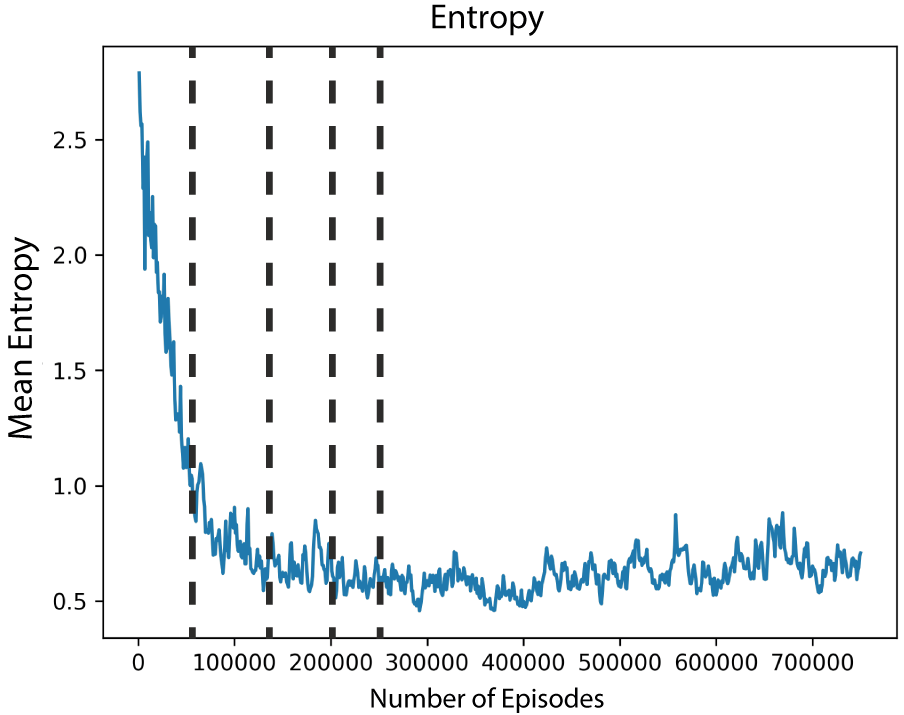} &
		\includegraphics[width=0.32\textwidth]{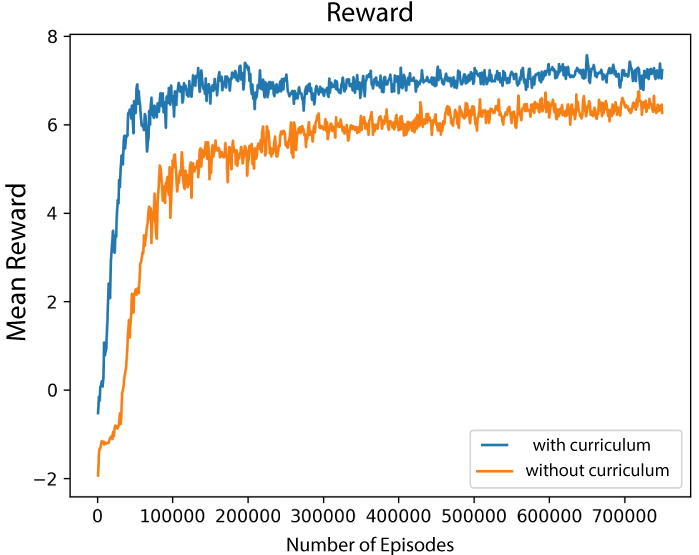}
	\end{tabularx}
	\caption{Plots showing metrics during the training phase for the 
		warrior class as a function of the number of episodes.
		From left to right: the evolution of the mean reward, 
		the evolution of the entropy, and the difference between
		the training with and without curriculum. The dashed vertical lines 
		on the plots delineate the different curriculum phases.
		\label{plots}
	}
\end{figure*}

\section{Implementation}
\label{sec:implementation}
In this chapter we describe how the DeepCrawl policy networks were
trained. We used two type of technologies to build both the DRL system
and the game:

\minisection{Tensorforce.}
Tensorforce~\cite{kuhnle17,lift-tensorforce} is an open-source DRL framework built on top of Google's TensorFlow framework, with an
emphasis on modular, flexible library design and straightforward
usability for applications in research and practice.
Tensorforce is agnostic to the application context or simulation, but offers an expressive state- and action-space specification API. In particular, it supports and facilitates working with multiple state components, like our global/local map plus property vector, via a generic network configuration interface which is not restricted to simple sequential architectures only. Moreover, the fact that Tensorforce implements the entire RL logic, including control flow, in portable TensorFlow computation graphs makes it possible to export and deploy the model in other programming languages, like C\# as described in the next section.

\minisection{Unity and Unity ML-Agents.}
The DeepCrawl prototype was developed with Unity~\cite{unity} and
Unity Machine Learning Agents Toolkit (Unity
ML-Agents)~\cite{juliani18}, that is an open source plugin available for the game engine that enables
games and simulations to serve as environments for training
intelligent agents. This framework allows external Python libraries to
interact with the game code and provides the ability to use
pre-trained graph directly within the game build thanks to
the TensorFlowSharp plugin~\cite{tensorflowsharp}.

\begin{table*}[ht]
	\caption{Results of the SEQ questionnaire. Players
		answered with a value between 1 (strongly disagree)
		and 7 (strongly agree).}
	\label{seq_table}
	\vskip 0.15in
	\begin{center}
		\begin{small}
			\begin{sc}
				\scalebox{0.84}{
					\begin{tabular}{c|l|c|c|c|c}
						\toprule
						N° & Question & \multicolumn{1}{c|}{$\mu$} & Mo & Md &\multicolumn{1}{c}{$\sigma$} \\ 
						\midrule
						1  & Would you feel able to get to the end of the game? & 5.54 & 6 & 6 & 1.03 \\
						2  & As the level increases, have the enemies seemed too strong? & 4.63 & 4 & 5 & 0.67 \\
						3  & Do you think that the enemies are smart? & 5.72 & 5 & 6 & 0.78 \\
						4  & Do you think that the enemies follow a strategy? & 6.18 & 6 & 6 & 0.40 \\
						5  & Do you think that the enemies do counterintuitive moves? & 2.00 & 2 & 2 & 0.63 \\
						6  & Do the different classes of enemies have the same behavior? & 1.27 & 1 & 1 & 0.46 \\
						7  & Are the meaning of the icons and writing understandable? & 5.72 & 6 & 6 & 1.67 \\
						8  & Are the information given by the User Interface clear and enough? & 5.54 & 6 & 6 & 1.21 \\
						9  & Are the level too big and chaotic? & 2.00 & 1 & 2 & 1.34 \\
						10 & Are the objects in the map clearly visible? & 5.81 & 7 & 7 & 1.66 \\
						11 & Do you think that is useful to read the enemy's characteristics? & 6.90 & 7 & 7 & 0.30 \\
						12 & How much is important to have a good strategy? & 6.90 & 7 & 7 & 0.30 \\
						13 & Give a general value to enemy abilities compared to other Roguelike games & 6.00 & 6 & 6 & 0.77 \\
						14 & Is the game enjoyable and fun? & 5.80 & 6 & 6 & 0.87 \\
						15 & Does the application have bugs? & 1.09 & 1 & 1 & 0.30 \\ 
						\bottomrule
				\end{tabular}}
			\end{sc}
		\end{small}
	\end{center}
	\vskip -0.1in
\end{table*}

\subsection{Training setup}
To create agents able to manage all possible situations that can occur
when playing against a human player, a certain degress of randomness
is required in the procedurally-generated environments: the shape and
the orientation of the map, as well as the number of impassable and
collectible objects and their positions are random; the initial
position of the player and the agent is random; and the initial
equipment of both the agent and the player is random.


In preliminary experiments we noticed that agents learned very slowly,
and so to aid the training and overcome the problem of the sparse
reward function, we use curriculum learning~\cite{bengio09} with
phases shown in figure~\ref{curriculum_timeline}. This technique lets
the agent gradually learn the best moves to obtain victory: for
instance, in the first phase it is very easy to win the game, as the
enemy has only 1 HP and only one attack is needed to defeat it; in
this way the model can learn to reach its objective without worrying
too much about other variables.  As training proceeds, the environment
becomes more and more difficult to solve, and the ``greedy'' strategy
will no longer suffice: the agent HP will vary within a range of
values, and the enemy will be more difficult to defeat, so it must
learn how to use the collectible items correctly and which attack is
the best for every situation. In the final phases loot can be
difficult to find and the HP, of both agent enemy, can be within a large
range of values: the system must develop a high level of strategy to
reach the end of the game with the highest score possible.

\minisection{Agents opponent.}
The behavior of the enemies agents are pitted against is \textit{of great
importance}. To satisfy requirements defined in
section~\ref{desiderata}, during training the agents fight against 
an opponent that always makes \textit{random moves}.
In this way, the agent sees all the possible
actions that a user might perform, and at the same time it can be
trained against a limited enemy with respect of human
capabilities. This makes the agent beatable in the long run, but still
capable of offering a tough and unpredictable challenge to the human player.



\subsection{Training results}
\label{train_par}

The intrinsic characteristic values for NPCs must be chosen before
training. These parameters are not observed by the system, but offer an
easy method to create different types of agents. Changing the agent's
ATK, DEF or DEX obliges that agent to extrapolate the best strategy
based on its own characteristics. For DeepCrawl we trained three
different combinations:
\begin{itemize}
	\itemsep0em
	\item \textbf{Archer}: ATK = 0, DEX = 4 and DEF = 3;
	\item \textbf{Warrior}: ATK = 4, DEX = 0 and DEF = 3; and
	\item \textbf{Ranger}: ATK = 3, DEX = 3 and DEF = 3.
\end{itemize}
For simplicity, the opponent has always the same characteristics: ATK =
3, DEX = 3 and DEF = 3.

To evaluate training progress and quality, we performed some
quantitative analysis of the evolution of agent policies. In
figure~\ref{plots} we show the progression of the mean reward
and entropy for the warrior class as a function of the number of
training episodes. The other two types of agents follow the same
trend. In the same figure we show the difference between
training with and without curriculum learning. Without curriculum, the
agent learns much slower compared to multi-phase curriculum training.
With a curriculum the agent achieves a significantly higher
average reward at the end of training.

\subsection{PPO and hyperparameters}
To optimize the policy networks we used the PPO
algorithm.  One agent rollout is made of 10
episodes, each of which lasts at most 100 steps, and it may end either
achieving success (i.e. agent victory), a failure (i.e. agent death)
or reaching the maximum steps limit. At the end of 10 episodes, the
system updates its weights with the episodes just experienced.  PPO is
an Actor-Critic algorithm with two functions that must be learned: the
policy and the baseline. The latter has the goal of a normal state
value function and, in this case, has the exactly same structure as
the policy network show in figure \ref{architecture}.

Most of the remaining hyper-parameters values were chosen after many
preliminary experiments made with different configurations: the policy
learning rate $lr_p = 10^{-6}$, the baseline learning rate
$lr_b = 10^{-4}$, the agent exploration rate $\epsilon = 0.2$, and the
discount factor $\gamma = 0.99$.

\section{Playability evaluation}
\label{sec:evaluation} 

To evaluate the DeepCrawl prototype with respect to our desiderata, we
conducted playability test as a form of qualitative analysis. The
tests were administered to 10 candidates, all passionate videogamers
with knowledge of the domain; each played DeepCrawl for sessions
lasting about 60 minutes. Then, each player was asked to answer a
Single Ease Question (SEQ) questionnaire. All the questions were
designed to understand if the requirements laid out in
section~\ref{desiderata} had been met and to evaluate the general
quality of DeepCrawl. Table~\ref{seq_table} summarizes the results.

We reconsider here each of the main requirements we discussed above in
section~\ref{desiderata} in light of the player responses:
\begin{itemize}
	\itemsep0em
	\item \textbf{Credibility}: as shown by questions 3, 4, and 5, the
	agents defined with this model offer a tough challenge to players;
	the testers perceived the enemies as intelligent agents that follow
	a specific strategy based on their properties.
	
	\item \textbf{Imperfection}: at the same time, questions 1 and 2
	demonstrate that players are confident they can finish the game with
	the proper attention and time. So, the agents we have trained seem
	far from being superhuman, but they rather offer the right amount of
	challenge and result in a fun gameplay experience (question 14).
	
	\item \textbf{Prior-free}: questions 5 and 12 show that, even with a
	highly sparse reward, the model is able to learn a strategy without
	requiring developers to define specific behaviors. Moreover,
	question 13 indicates that the agents implemented using DRL are
	comparable to others in other Roguelikes, if not better.
	
	\item \textbf{Variety}: the testers stated that the differences
	between the behaviors of the distinct types of agents were very
	evident, as shown by question 6. This gameplay element was much
	appreciated as it increased the level of variety and fun of the
	game, and improved the general quality of DeepCrawl.
\end{itemize}

\section{Conclusions}
\label{sec:conclusion}

In this we presented a new DRL framework for development of NPC agents
in video games. To demonstrate the potential of DRL in video game
production, we designed and implemented a new Roguelike called
DeepCrawl that uses the model defined in this article with excellent
results. The current versions of the agents work very well, and the
model supports numerous agents types only by changing a few
parameters before starting training. We feel that DRL brings many
advantages to the table commonly used techniques like
finite state machines or behavior trees.

The recent and highly-publicized successes of DRL in
mimicking or even surpassing human-level play in games like
DOTA have not yet been translated into effective tools for use in
developing game AI. The DeepCrawl prototype is a step in this
direction and shows that DRL can be used to develop credible -- yet
imperfect -- agents that are prior-free and offer variety to gameplay
in turn-based strategy games. We feel that DRL, with
some more work towards rendering training scalable and flexible, can
offer great benefits over classical, hand-crafted agent design that
dominates the industry.

\bibliographystyle{aaai}
\bibliography{main}

\end{document}